\documentclass{article} % For LaTeX2e
\usepackage{collas2025_conference,times}
\usepackage{easyReview}
\usepackage{algorithm}

\usepackage{booktabs}
\usepackage{latexsym}
\usepackage{amssymb}
\usepackage{amsmath}
\usepackage{amsthm}
\usepackage{algpseudocode}
\usepackage{subcaption}
\usepackage{graphicx}
\usepackage{color}

\usepackage{newfloat}
\usepackage{listings}
\usepackage{amsmath,amsfonts,bm}

\def\eqref#1{equation~\ref{#1}}
\def\1{\bm{1}}

\DeclareMathAlphabet{\mathsfit}{\encodingdefault}{\sfdefault}{m}{sl}
\SetMathAlphabet{\mathsfit}{bold}{\encodingdefault}{\sfdefault}{bx}{n}

\usepackage{hyperref}
\hypersetup{
    colorlinks=true,
    linkcolor=red,
    filecolor=magenta,
    urlcolor=blue,
    citecolor=purple,
    pdftitle={Overleaf Example},
    pdfpagemode=FullScreen,
    }

\title{On Supernet Transfer Learning for Effective Task Adaptation}

\author{Prabhant Singh, Joaquin Vanschoren \\
Eindhoven University of Technology\\
Netherlands \\
\texttt{\{p.singh,j.vanschoren\}@tue.nl} \\
}

\collasfinalcopy % Uncomment for camera-ready version, but NOT for submission.

\begin{document}

\maketitle
\begin{abstract}
Neural Architecture Search (NAS) methods have been shown to outperform hand-designed models and help to democratize AI. However, NAS methods often start from scratch with each new task, making them computationally expensive and limiting their applicability. Transfer learning is a practical alternative with the rise of ever-larger pretrained models. However, it is also bound to the architecture of the pretrained model, which inhibits proper adaptation of the architecture to different tasks, leading to suboptimal (and excessively large) models.
We address both challenges at once by introducing a novel and practical method to \textit{transfer supernets}, which parameterize \textit{both} weight and architecture priors, and efficiently finetune both to new tasks.
This enables `\textit{supernet transfer learning}' as a replacement for traditional transfer learning that also finetunes model architectures to new tasks.
Through extensive experiments across multiple image classification tasks, we demonstrate that supernet transfer learning does not only drastically speed up the discovery of optimal models (3 to 5 times faster on average), but will also find better models than running NAS from scratch. The added model flexibility also increases the robustness of transfer learning, yielding positive transfer to even very different target datasets. Finally, we find that multi-dataset supernet pretraining provides a powerful and practical approach to transfer learning.
\end{abstract}    
\section{Introduction}
Neural architecture search (NAS) has repeatedly demonstrated the ability to automatically design state-of-the-art neural architectures, and reduce the time and effort that goes into designing neural networks from scratch ~\citep{nassurvey, firstnas}. However, this search for architectures can be computationally very expensive. Although there have been many improvements, such as smart search space design, meta-learning \citep{vanschoren2018metalearningsurvey}, zero-cost proxies~\citep{zerocost}, and faster optimization algorithms~\citep{nao}, most methods still start from scratch with every new task. There has been limited research into combining transfer learning and NAS into a coherent whole, transferring both model weights and architectures while allowing both to be finetuned to new tasks.

\begin{figure}
    \centering
    \includegraphics[width=0.8\columnwidth]{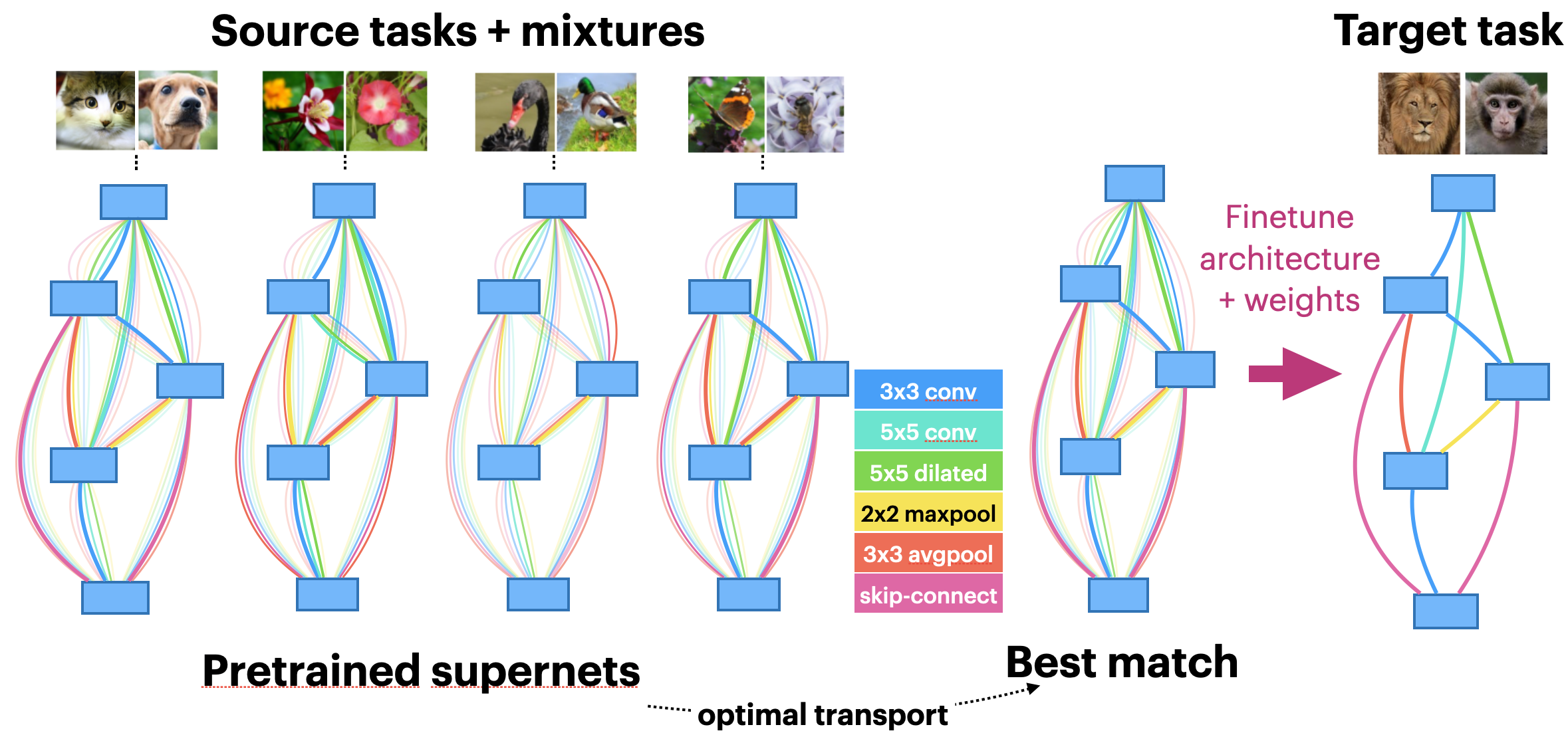}
    \caption{Starting from a zoo of supernets $\theta(D_i)$ pretrained on a large collection of source tasks $\mathcal{D}_{meta}$, and a target task $\mathcal{D}_{target}$, we select the best match using Optimal Transport. We then continue to train the pretrained supernet using SmoothDARTS, finetuning both the architecture and weights to any given objective, and then derive the final architecture and finetune the remaining model weights further to obtain the final model.}
    \label{fig:transferfig}
    %\vspace{10mm}
\end{figure}

One of the most studied frameworks for NAS is Differentiable architecture search (DARTS; ~\citet{darts}) which trains an over-parameterized architecture (a \textit{supernet}) that parameterizes both the architecture configuration and the model weights. Using bilevel optimization and weight sharing, this method can optimize model and architecture parameters simultaneously, which is drastically more efficient. Major further improvements have been made to make this approach more robust, stable, efficient, and generalizable, combined in methods such as SmoothDARTS \citep{smoothdarts}. 

% Adjust mm as needed
In DARTS, supernets are trained on a given dataset and then discretized into a final model by keeping only the most important operations (layers), at which point the supernets are discarded. However, these supernets have learned architectural biases (inherently useful operations and structures) and representations (features) that are likely beneficial for related tasks. In this work, we explore how we can effectively select and transfer supernets pretrained on prior tasks, and fine-tune them to new (target) tasks. As illustrated in Figure \ref{fig:transferfig}, we keep a memory (or \textit{zoo}) of pretrained supernets, each pretrained on different image classification source tasks. The graphs represent a small part (a \textit{cell}) of the supernets, where the boxes are feature maps and the colored edges are possible operations (e.g. a 3x3 convolution or 2x2 maxpooling), and the thickness of the edges represents the learned weight of each operation at each location in the architecture, which we call the architecture weights. The model weights themselves are not shown here but are also stored in the supernet for every possible operation (layer). 

For each source task, the pretrained supernet weights will be different, but likely similar for related tasks. As in source model selection~\citep{transferwhich}, when given a new (target) task, we select the best supernet from our zoo and then finetune it to the new tasks. In particular, we employ Optimal Transport measures to select source tasks that are similar to the target tasks and transfer the corresponding pretrained supernet, assuming that it will transfer best to the new task since it was pretrained on similar data. In addition, for situations where it may be impractical to store zoos of pretrained supernets, we also use multi-dataset pretraining techniques to pretrain supernets on a wide range of source tasks at once.

Supernet training is expensive for DARTS-like frameworks. Hence, by effectively transferring supernets we can significantly speed up NAS. Note that this is very different from transferring only the model weights as is usually done in traditional transfer learning: since we also transfer the architecture weights, we can fine-tune the architecture itself by warm-starting DARTS with pretrained model and architecture weights. As we are not confined to a fixed architecture, we can also finetune the architecture towards additional objectives (e.g. balancing model size and performance).

%By transferring the supernet, we effectively transfer the state of the architecture search process, before the supernet is collapsed into a discrete neural architecture, and then continue the architecture search toward a new target task.
% Joaquin: skipped this because I think it's clear by now. Also, I don't want people to think we're transferring the optimizer state.

A valid concern would be that pretrained supernets may be too biased to use as starting points. However, we show that by using a diverse zoo of pretrained supernets, using pre-existing optimal transport-based metrics, and appropriate NAS finetuning, we achieve positive transfer to new tasks, and find models that are better than running DARTS from scratch.

%When one has multiple tasks and pre-trained networks available to transfer from then the selection of the best source dataset or model becomes a difficult problem, since this selection can result in positive transfer as well as negative transfer (in which the fine-tuned model performs worse than the one trained from scratch). To achieve robust task selection for transfer learning in pre-trained supernets we use optimal transport-based metrics in this paper. Our study shows the existence of negative and positive transfer in DARTS, and that we can achieve robust positive transfer using Optimal Transport-based dataset selection for supernet finetuning.
\par
%Through this work, we aim to contribute to the understudied area of transfer learning in Neural Architecture Search. 
In summary, we make the following contributions:
\begin{enumerate}
    \item We present a novel setup for supernet transfer learning that transfers both architecture and representation priors. We explore both supernet selection based on optimal transport and large-scale supernet pretraining. 
    \item We present a comprehensive study across dozens of vision tasks to evaluate our method and empirically show the effectiveness of supernet transfer learning, both in terms of speed (3 to 5 times faster on average), and performance (often outperforming other NAS techniques).
    \item We provide new insights into the practical aspects of supernet transfer learning, especially that multi-dataset supernet pretaining leads to faster convergence and better generalization.
    \end{enumerate}

\section{Related Work}
To the best of our knowledge, no prior studies have proposed and evaluated transfer learning using DARTS supernets, or optimal selection of proxy sets for supernet training. Several works have explored meta-learning and transfer learning on the final (discretized) architectures obtained using Neural Architecture Search (NAS). We consider these works interesting and academically related to ours but not directly comparable since the supernets are not kept or transferred. NASTransfer~\citep{nasnanlysisi} presented a first analysis in this area with several NAS methods including DARTS, ENAS~\citep{enas}, and NAO~\citep{nao}. This study analyzed the transferability of the obtained architectures but didn't actually transfer the supernets themselves. MetaNAS~\citep{metanas} proposed a gradient-based approach to meta-learn architecture weights and allow fast finetuning to related tasks. It combines DARTS with REPTILE~\citep{reptile} to make NAS possible for few-shot learning scenarios. However, as a meta-learning technique, it is designed to specialize to a distribution of very similar tasks, not transfer to potentially different tasks. 

Meta Dataset-to-Architecture (MetaD2A; \citet{metad2a}) directly generates architectures using meta-learning, based on meta-features describing the target dataset. It doesn't transfer supernets or any model weights.
Transferable NAS (TNAS;~\citep{transfernas}) is a Bayesian Optimization (BO) technique based on a surrogate model that, given a dataset embedding, predicts the utility of different architectures. While it can recommend architectures for a target dataset, it does not transfer model weights or allow us to warmstart NAS. However, it only transfers model weights, not the architecture biases present in the supernets. As such, it is still bound to previously existing models and cannot finetune the architectures themselves towards new objectives.

\section{Background}

% \subsection{Differentiable architecture search}
DARTS is one of the most widely researched NAS frameworks. Based on the observation that state-of-the-art neural architectures consist of blocks (cells) that are stacked together, e.g. 'normal' and 'reduction' blocks in CNN architectures, it aims to learn the optimal architecture for these cells. In a cell, $N$ nodes are arranged as a Directed Acyclic Graph (DAG), as shown in Figure 1. Each (blue) node $x^{(i)}$ represents a latent feature map, and each colored edge $(i, j)$ is a specific operation $o(i,j)$. Since operations are discrete in nature, a relaxation technique is used that gives each operation a learnable architecture weight $\alpha_{o}^{(i,j)}$ (represented as edge thickness) and the outputs of all possible operations are combined using a softmax over learned architecture weights to allow backpropagation over these weights:

\begin{equation}
    \bar{o}^{(i,j)}(x) = \sum_{o \in \mathcal{O}} \frac{\exp(\alpha_{o}^{(i,j)})}{\sum_{o' \in \mathcal{O}} \exp(\alpha_{o'}^{(i,j)})} o(x),
\label{eq: 1}
\end{equation}

In Equation \ref{eq: 1}, $\mathcal{O}$ is the candidate operation corpus and $\alpha_{o}^{(i,j)}$ is the corresponding architecture weight for operation $o$ on the edge $(i,j)$. The architecture search is thus relaxed to learning a continuous architecture weight vector \textbf{ $A = [\alpha^{(i,j)}]$}, resulting in a bilevel optimization objective:

\begin{equation}
\begin{split}
    \min_{A} \mathcal{L}_{\text{val}}(w^*(A), A) \\
    \text{s.t.} \quad w^* = \underset{w}{\mathrm{arg\,min}} \, &\mathcal{L}_{\text{train}}(w, A).
\end{split}
\label{eq: 2}
\end{equation}

where $w$ are the model weights and $\mathcal{L}$ is the loss function. During training, DARTS performs one or more gradient descent steps to update $w$ using the training data and the current architecture weights $A$, and then performs one or more gradient descent steps to update $A$ using the validation data, in an interleaved fashion, thus jointly learning the architecture and model weights. The supernet is the network consisting of all possible operations, their architecture weights, and their model weights. When converged, the supernet is discretized by choosing the operation for each edge with the largest architecture weight, and this final architecture is then finetuned again to obtain the final model weights. 

SmoothDARTS additionally employs perturbation-based regularization techniques, such as random smoothing and adversarial training, to smooth the validation loss landscape, enhancing model stability and generalizability. This regularization implicitly regulates the Hessian norm of the validation loss, leading to more robust architectures. It also builds on RobustDARTS~\citep{robustdarts} which minimizes the performance drop during discretization and improves generalization.

\begin{algorithm}
    \caption{Supernet pretraining}
        \begin{flushleft} 
        \textbf{Inputs:} \\
        The source dataset collection: $\mathcal{D}_{\text{meta}}$ \\
    An untrained supernet $\theta$ with model weights $w$ and architecture weights $A$: $\theta = (w, A)$ \\
        \textbf{Outputs:} \\
            The list of pre-trained supernets: $\mathbf{\Theta}_{\text{meta}}$\\

        \end{flushleft}

    \begin{algorithmic}[1]
     \State Initialize $\mathbf{\Theta_{meta}}$ to []
     \For{$D_i$ in $\mathcal{D}_{\text{meta}}$}
    \State Randomly initialize $\theta_i$
    \While{not converged}
	\State Update $A$ by descending $\nabla_A \mathcal{L}_{val}(w,A)$
	\State Update $w$ by descending $\nabla_w \mathcal{L}_{train}(w, A)$ 
    \EndWhile
    \State Append $\theta_i$ to $\mathbf{\Theta}_{\text{meta}}$
        \EndFor
    \end{algorithmic}
    \label{algo:metatraining}
\end{algorithm}

\section{Supernet Transfer Learning}
\begin{algorithm}[t]
    \caption{Supernet Transfer Learning}
    \label{algo:findtransfer}
    \begin{flushleft} 
        \textbf{Inputs:} \\
        The target dataset: $D_{\text{target}}$ \\
        The source dataset collection: $\mathcal{D}_{\text{meta}}$ \\
        The list of pre-trained supernets: $\mathbf{\Theta}_{\text{meta}} = (\mathbf{w}_{\text{meta}}, \boldsymbol{A}_{\text{meta}})$ \\
        An embedding network to embed all datasets: $\phi: D \rightarrow \mathbb{R}^k$
    \end{flushleft}
    \begin{algorithmic}[1]
        \State Initialize $\mathbf{d}_{\text{ot}}$ to $[]$
        \For{$D_i$ in $\mathcal{D}_{\text{meta}}$}
            \State $d_{\text{ot}_i} \gets \text{OT}(\phi(D_{\text{target}}), \phi(D_i))$ \Comment{Distance calc.}
            \State Append $d_{\text{ot}_i}$ to $\mathbf{d}_{\text{ot}}$
        \EndFor
        \State $s \gets \text{argmax}(1 - \mathbf{d}_{ot})$ \Comment{Index of most similar dataset}
        \State $w_s,A_s \gets  \mathbf{w}_{\text{meta}}[s], \boldsymbol{A}_{\text{meta}}[s]$ \Comment{Get supernet ($w_s,A_s$)}
        \State $w_t,A_t \gets w_s,A_s$ \Comment{Transfer weights}
        \State Warm-start DARTS with $w_t,A_t$ 
        \While{not converged}
        \State Finetune $w_t,A_t$ based on equation \ref{eq: 2}
        \EndWhile
        \State Derive final architecture $A_T$ based on the learned $A_t$
        \State Finetune the remaining model weights $w_t(A_T)$ on $D_{Target}$.
    \end{algorithmic}
\end{algorithm}
\subsection{OT based Supernet Transfer}
We hypothesize that when two datasets share a high degree of similarity, i.e. they have similar data distributions, the learned features and architectural biases of a supernet pre-trained on one dataset can be effectively transferred to the other. Quantifying this similarity is a challenging task, however, particularly when dealing with complex data distributions, such as those found in image datasets, rather than individual data points. To address this, Optimal Transport and the Wasserstein distance can be repurposed to measure the "distance" between two probability distributions. We can then assume that the smaller the distance, the greater the similarity between the datasets.

% Optimal Transport (OT) can be efficiently solved by the Sinkhorn algorithm~\cite{Cuturi2013SinkhornDL}, low-rank linear-time approximations~\cite{lowrank}, and neural network-based solutions~\cite{Amos2022MetaOT}. Entropic regularized OT has been used in domain adaptation~\cite{DAOT, jdot}, and recent works like OTDD~\cite{otdd} and OTCE~\cite{otce} apply OT to find transfer learning metrics between image datasets.

First, we build a collection of labeled datasets $\mathcal{D}_{meta}$ which will serve as a collection of source datasets $\mathcal{D}_{\text{meta}} = \{D_1, D_2, \ldots, D_n\}$. Next, we use SmoothDARTS to pretrain a supernet on each of them, yielding a collection of pre-trained supernets $\Theta_{meta} = \{\theta_1, \ldots, \theta_n\}$ as described in Algorithm \ref{algo:metatraining}. Given a target dataset $D_{Target} \notin \mathcal{D}_{meta}$, we aim to find the most similar source dataset $D_s$ and the corresponding pretrained supernet $\theta_s$.

To find the OT distance between these datasets, we first need to embed these datasets within a metric space with an embedding function $\phi$, because OT can only be applied to distributions with the same dimensionality. $\phi$ can be any neural network that can be used to embed both the datasets in  $\mathcal{D}_{meta}$ and $D_{Target}$.  

\begin{equation}
    d_{ot_i} = OT(\phi(D_{target}, D_{source(i)})
\end{equation}
In this work we use Bures-Wasserstein distance by \cite{otdd}.

For our specific DARTS method, we use SmoothDARTS because it is one of the most stable versions of DARTS and it does not use any zero-cost proxies or hacks, which allows us to better analyze the effect of transfer learning. SmoothDARTS also builds on multiple other DARTS-like frameworks that are widely accepted by the community. Moreover, SmoothDARTS minimizes the performance gap incurred by discretizing the final supernet into a final model, leading to better generalization. However, other DARTS methods like PC-DARTS~\citep{PC-DARTS} and DRNAS~\citep{drnas} could certainly be chosen instead. 

In the remainder of this paper, we will evaluate two variants of our supernet transfer approach:
\begin{itemize}
    \item \textbf{OT Transfer}, which transfers the supernet pretrained on a single source dataset, selected by OT($d_{ot}$).
    \item \textbf{Multi-Dataset Transfer}, which transfers the supernet pretrained on a mixture of all source datasets combined. This is shown in Algorithm \ref{algo:findmix}.
\end{itemize}

Both can be valuable depending on the application and the availability of prior datasets and pretrained supernets. 

\begin{algorithm}[tb]
    \caption{Multi-dataset Supernet Transfer}
    \label{algo:findmix}
    \begin{flushleft} 
        \textbf{Inputs:} \\
        The target dataset: $D_{\text{target}}$ \\
        The source dataset collection: $\mathcal{D}_{\text{meta}}$ \\
        A supernet pretrained on all of $\mathcal{D}_{\text{meta}}$: $\Theta_{\text{mix}} = (w_{\text{mix}}, A_{\text{mix}})$ \\
    \end{flushleft}
    \begin{algorithmic}[1]
        \State Warm-start DARTS with $w_{mix},A_{mix}$ 
        \While{not converged}
        \State Finetune $w_{mix},A_{mix}$ based on equation \ref{eq: 2}
        \EndWhile
        \State Derive final architecture $A_T$ based on the learned $A_{mix}$
        \State Finetune $w_{mix}(A_T)$ on $D_{Target}$.
        % \State Discretize $\alpha_{mix}$
    \end{algorithmic}
\end{algorithm}

\section{Experimental analysis}
In this section, we present a comprehensive study across dozens of vision tasks to evaluate our method and empirically analyze the effectiveness of supernet transfer learning. We specify the datasets used, the architecture search space, as well as the learning objective and metrics. Next, we break down our analysis into 5 questions and answer them one by one. Additional details and results, including runtimes and examples of the created architectures, are provided in the Supplementary material. The code for our method and experiments, as well the pretrained supernets, are available on GitHub.
 \begin{figure}
     \centering
    \includegraphics[width=0.75\columnwidth]{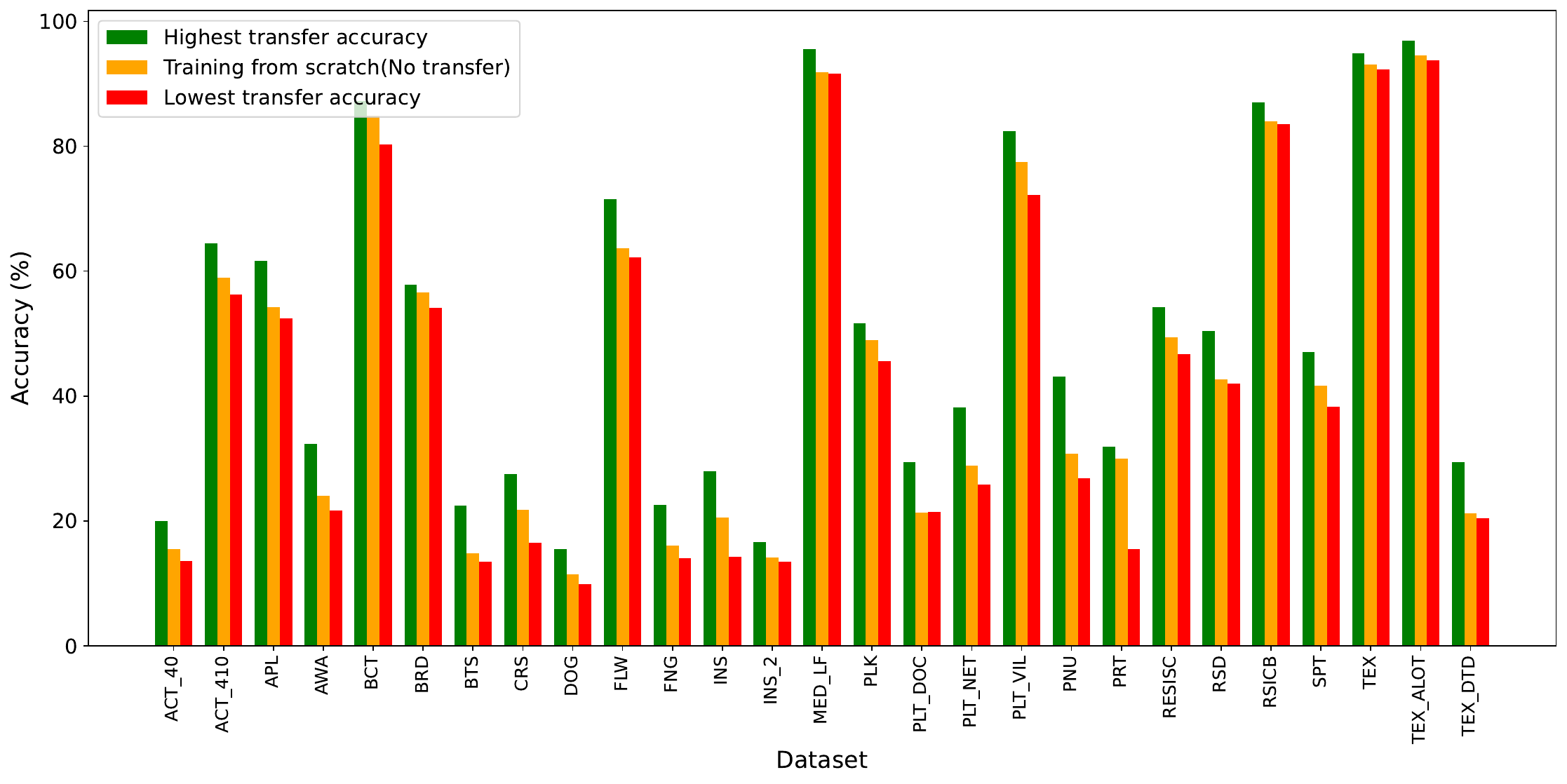}
    \caption{Both positive and negative transfer can occur depending on the selected source datasets}
    \label{fig:transfer_all}

 \end{figure}
     
    \vspace{2mm}
\textbf{Datasets: }
To build a rich collection of source datasets and supernets, we use datasets from the Meta-Album collection. Meta-Album~\citep{meta-album-2022} is an image classification meta-dataset, hosted on OpenML~\citep{OpenML},  designed to facilitate computer vision research, including transfer learning, continual learning, and meta-learning. A detailed list of all meta album datasets used is and their characteristics is present in Supplementary material.
\par

\textbf{Search Space: }
We use the \textit{S1} search space from SmoothDARTS~\citep{smoothdarts} as it is widely used in related work. We further adapt the SmoothDARTS settings to incorporate the Meta-album datasets.

\par
\textbf{Objective and Metrics: }
We aim to analyze the performance of Supernet Transfer using SmoothDARTS supernets, and learn from these findings how to create more efficient and robust NAS and transfer learning methods. Following ~\citet{nasnanlysisi} we compute top-1 classification accuracy of runs on baselines as a performance metric in our experiments. We also compute relative improvement for all baselines. Relative improvement can be described as $RI = \frac{ACC_a - ACC_b}{ACC_b} $ where $ACC_a$ is the accuracy of the method, and $ACC_b$ is the accuracy of the baseline. 

\par
\textbf{Runtimes: }
We ran  SmoothDARTS both from scratch and warm-started for 50 epochs to compare learning curves, although the warm-started runs often converged much sooner than that. Every SmoothDARTS run took 3-4 hours on average on a single NVIDIA A8000 GPU.

\par
\textbf{Experiments: }
We run several experiments in our analysis to answer these 5 questions:

\begin{enumerate}
    \item Is there any benefit in selecting the best source dataset in supernet transfer? Does the choice of source dataset lead to negative transfer (decrease in performance vs training from scratch) or positive transfers (increase in performance after transfer)?
    \item Are OT-based dataset distances effective in finding optimal pretrained supernets for transfer learning?
    \item How fast does supernet transfer converge? Is the size of the source dataset relevant to the performance of transfer learning?
    \item What is the impact of supernet transfer on convergence?
    \item Does warm-starting lead to better final models?
\end{enumerate}
\begin{table*}[h]
    \centering
\begin{tabular}{|c|c|c|c|}
\toprule
 \textbf{Dataset} &         \textbf{RI\_OT} &          \textbf{RI\_MDT} &       \textbf{RI\_Oracle} \\
\midrule
     WHOI-Plankton~\citep{whoiplankton} &          0.0059 & \textbf{0.0701} &          0.0558 \\
     Flowers~\citep{flowers} &          0.0046 & \textbf{0.1880} &          0.1233 \\
     100 Sports~\citep{100-sports} &          0.0214 & \textbf{0.2253} &          0.1299 \\
     Birds~\citep{birds} &          0.0050 & \textbf{0.0875} &          0.0210 \\
 Plant Village~\citep{plantvillage} &          0.0628 & \textbf{0.1154} &          0.0628 \\
     Textures~\citep{textures} &          0.0143 & \textbf{0.0285} &          0.0185 \\
     CARS~\citep{cars} &      \textbf{0.2653} &         0.2207 &     0.2653 \\
  RESISC45~\citep{RESISC} &         -0.0090 & \textbf{0.1483} &          0.0966 \\
  Stanford 40 Actions~\citep{pose} &          0.2893 & \textbf{0.4959} &          0.2893 \\
   SPIPOLL~\citep{Wu2019Insect} &          0.1806 & \textbf{0.3889} &          0.1806 \\
 PlantNet~\citep{garcin2021plntnetk} &          0.3264 & \textbf{0.4653} &          0.3264 \\
 Textures DTD\citep{cimpoi14describing} &          0.0854 & \textbf{0.6080} &          0.3920 \\
     Airplanes~\citep{wu_zhize_2019_3464319} &         -0.0219 & \textbf{0.1623} &          0.1360 \\
     PanNuke~\citep{gamper2019pannuke} &          0.0598 &          0.3932 & \textbf{0.4017} \\
     Stanford Dogs~\citep{KhoslaYaoJayadevaprakashFeiFei_FGVC2011} &          0.3564 & \textbf{0.5927} &          0.3564 \\
  Medicinal Leaf~\citep{s_j_2020} &         -0.0022 & \textbf{0.0588} &          0.0414 \\
   RSICB128~\citep{li2020RSI-CB} &          0.0185 & \textbf{0.0542} &          0.0357 \\
 MPII Human Pose~\citep{6909866}	 &          0.0643 & \textbf{0.0965} &          0.0936 \\
     Danish Fungi~\citep{picek2021danish} &          0.2000 & \textbf{0.8250} &          0.4125 \\
 PlantDoc~\citep{10.1145/3371158.3371196} &          0.3826 & \textbf{0.4087} &          0.3826 \\
Library of Textures (ALOT)~\citep{alot} &          0.0002 & \textbf{0.0370} &          0.0245 \\
     Animals with Attributes (AwA)~\citep{awa} &          0.3458 & \textbf{0.4292} &          0.3458 \\
     Insects~\citep{insects}
      &         -0.1028 &          0.3364 & \textbf{0.3575} \\
     RSD46~\citep{7827088} &          0.0679 & \textbf{0.3086} &          0.1821 \\
      Human Protein Atlas (Subcell)~\citep{thul2017subcellular} &         -0.0159 &         -0.0317 & \textbf{0.0635} \\
     MARVEL~\citep{MARVEL} &          0.0649 & \textbf{0.6494} &          0.5195 \\
\bottomrule
\end{tabular}
    \caption{Relative improvement of OT-transfer vs Multi-Dataset Transfer vs Oracle. Multi-Dataset Transfer almost always works significantly better than OT transfer and even the single-dataset Oracle, which establishes that more pretraining data does improve transferability.}
    \label{tab:lootable}
    \vspace{5mm}
\end{table*}

\subsection{Importance of pretrained supernet selection}\par
 To evaluate this question we transfer \textit{all} supernets $\Theta$ from $\mathcal{D}_{meta}$ for \textit{every} target dataset sampled from $\mathcal{D}_{meta}$ We first create a supernet dictionary $\mathbf{\Theta} = \{\Theta_1, \ldots, \Theta_n\}$ and run a simple grid search to find the optimal transfer dataset (the \textit{Oracle}). We also run SmoothDARTS \textit{from scratch} (without transfer), on all target datasets. The results are summarized in Figure \ref{fig:transfer_all}. This shows that both positive and negative transfer exists in supernet transfer for almost all target datasets, since the models found by warm-starting can be better or worse than those found by running SmoothDARTS from scratch. As such, the selection of good pretrained supernets is important for supernet transfer.

\subsection{Effectiveness of OT distances}
To answer this question we use Algorithm \ref{algo:findtransfer} to find the most similar dataset to the target dataset and then transfer the supernet to the target dataset and fine-tune it. The set of all measures OT distances is summarized in the heatmap in Figure \ref{fig: heatmap} in supplementary materials. We observe that `similar' datasets indeed seem related. For instance, PRT (Human Proteins) is considered similar to DIBaS (Bacteria), WHOI-Plankton (Plankton), and PNU (Human tissues), and all are microscopy images. We compare the top-1 accuracies of our method (OT Transfer) versus training SmoothDARTS and observe that we get a positive transfer in majority of the cases. (Results can be seen in Figure \ref{fig:transfer_otdd} in supplementary materials or Table \ref{tab:lootable} column RI\_OT).

% \begin{figure}[!htb]
%     \centering
%     \includegraphics[width=0.5\textwidth]{figures/loo.png}
%     \caption{Transfer learning from all datasets to the target dataset in a leave-one-out setting}
%     \label{fig:transfer_leave_one_out}
% \end{figure}

\begin{figure*}[t]
    \centering
    % First row with three subfigures
    \begin{subfigure}[htbp]{0.3\textwidth}
        \centering
        \includegraphics[width=\textwidth]{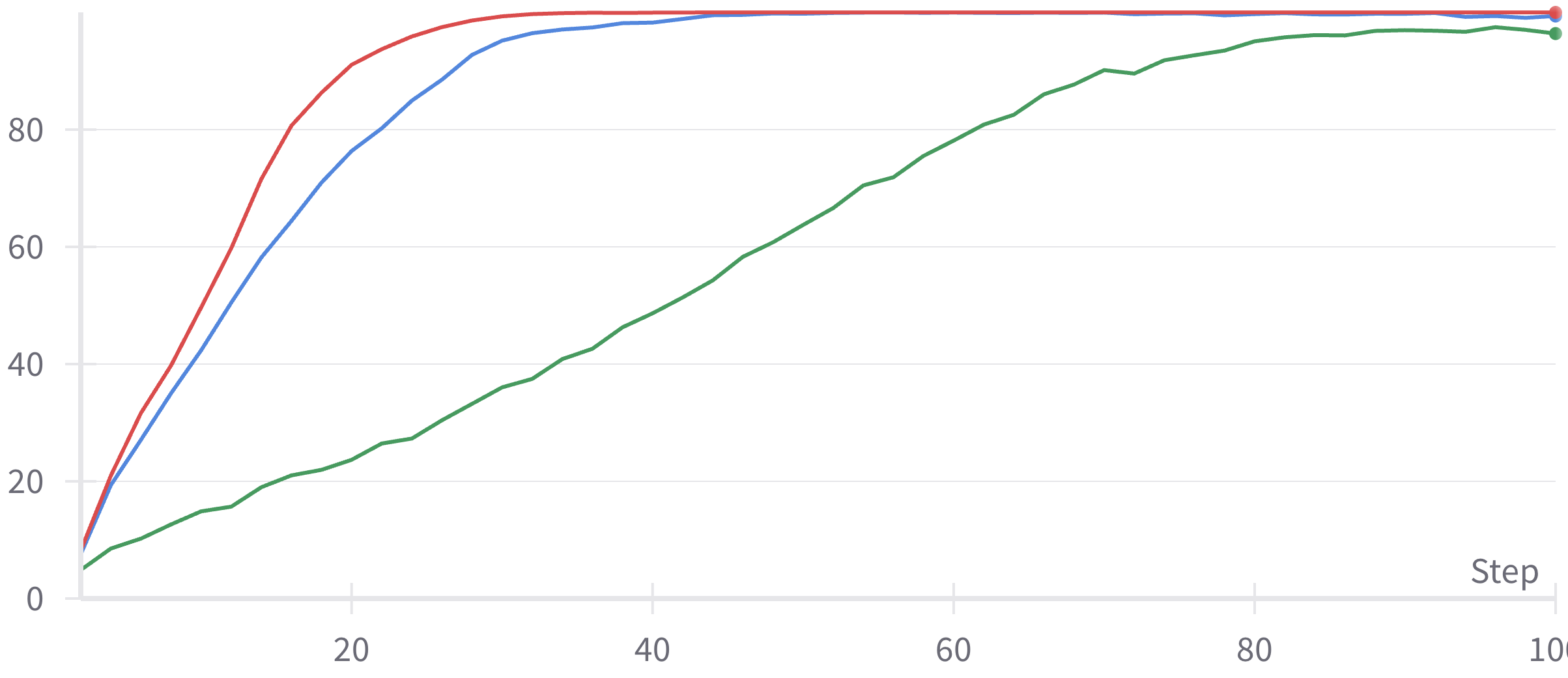}
        \caption{Stanford 40 Actions}
        \label{fig:sub1}
    \end{subfigure}
    \hfill
    \begin{subfigure}[htbp]{0.3\textwidth}
        \centering
        \includegraphics[width=\textwidth]{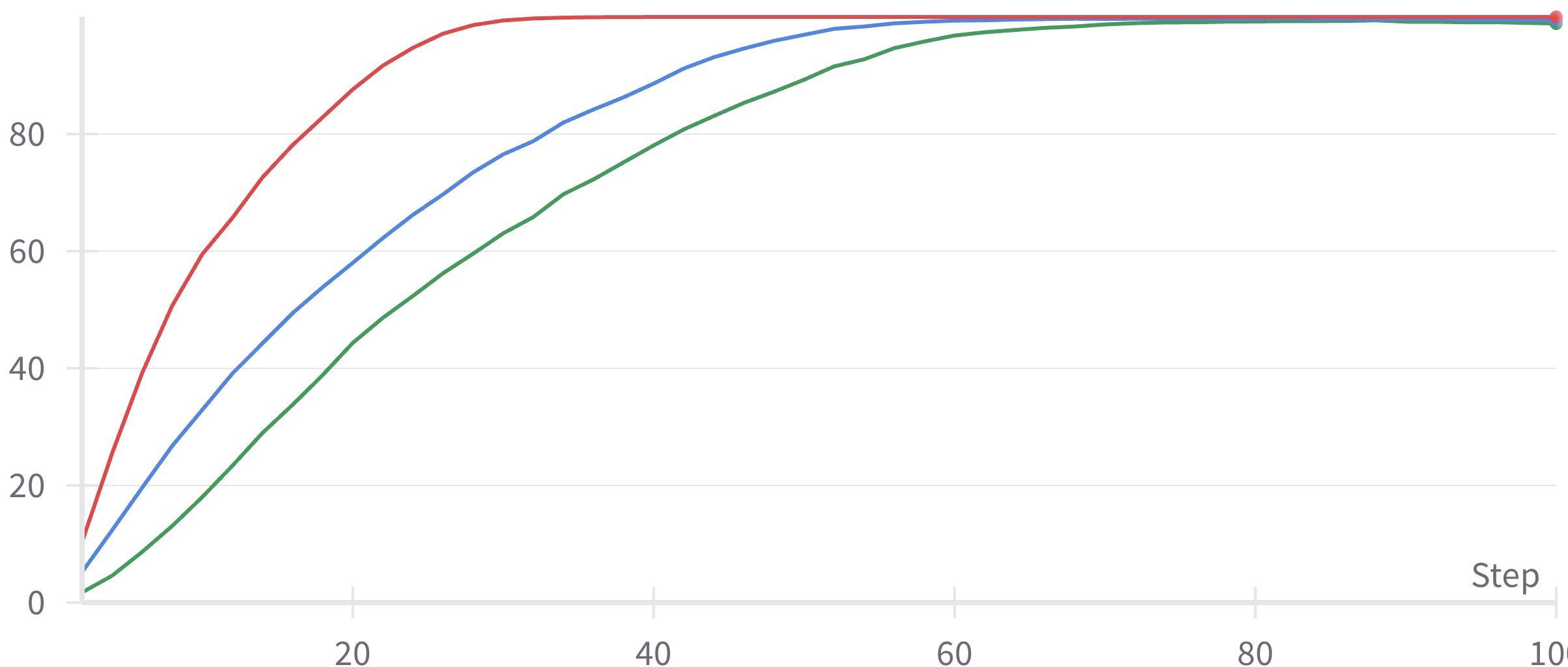}
        \caption{Birds}
        \label{fig:sub2}
    \end{subfigure}
    \hfill
    \begin{subfigure}[htbp]{0.3\textwidth}
        \centering
        \includegraphics[width=\textwidth]{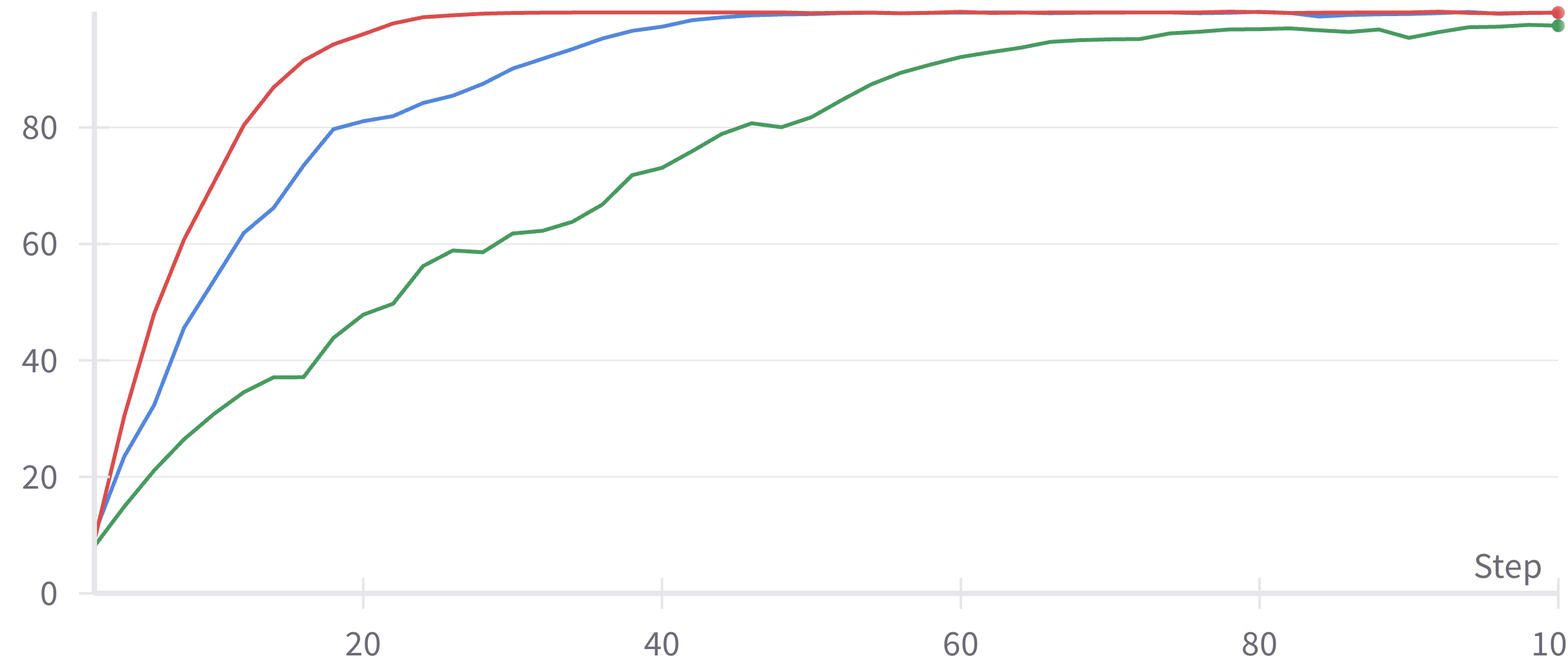}
        \caption{Airplances}
        \label{fig:sub3}
    \end{subfigure}
    % Add some vertical space before the next row
    \vspace{10pt} % Adjust the vertical space according to your needs

    % Second row with three subfigures
    \begin{subfigure}[htbp]{0.3\textwidth}
        \centering
        \includegraphics[width=\textwidth]{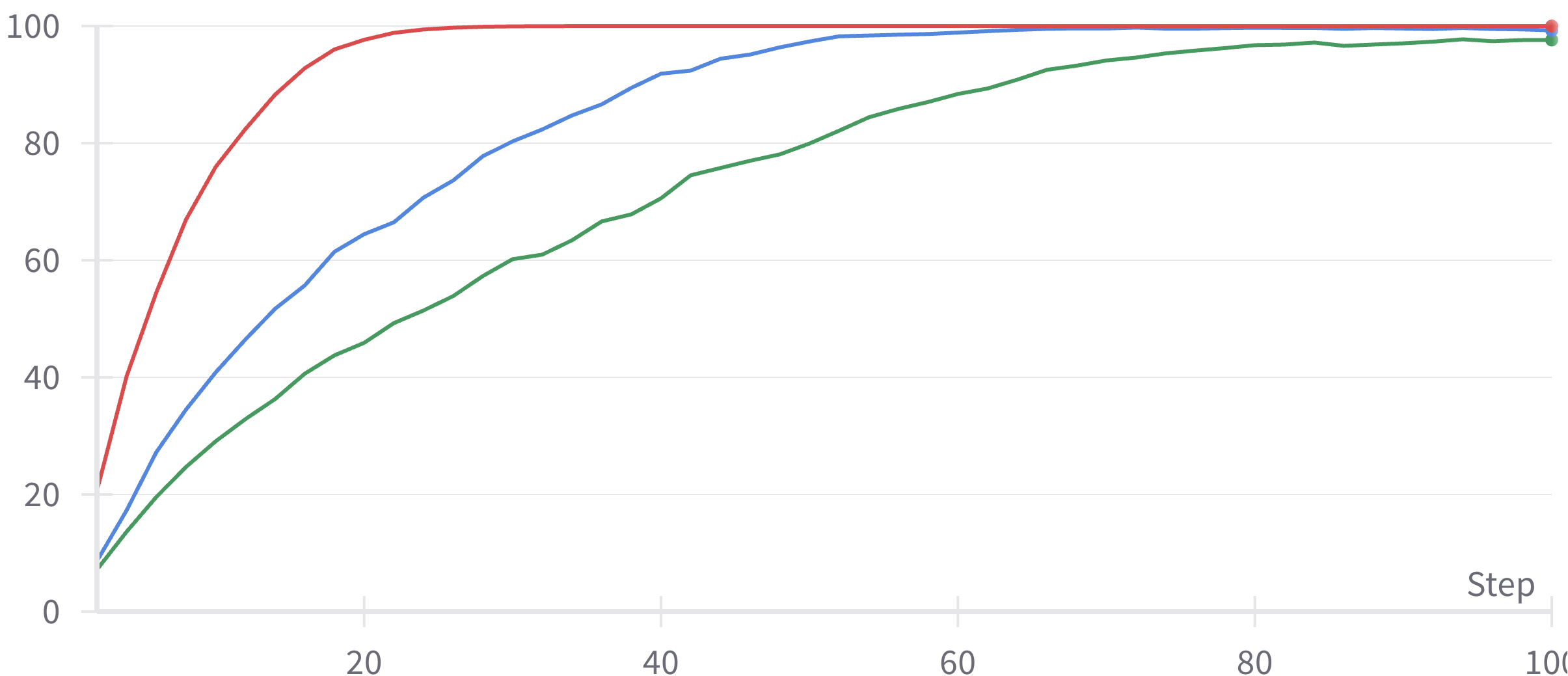}
        \caption{Flowers}
        \label{fig:sub4}
    \end{subfigure}
    \hfill
    \begin{subfigure}[htbp]{0.3\textwidth}
        \centering
        \includegraphics[width=\textwidth]{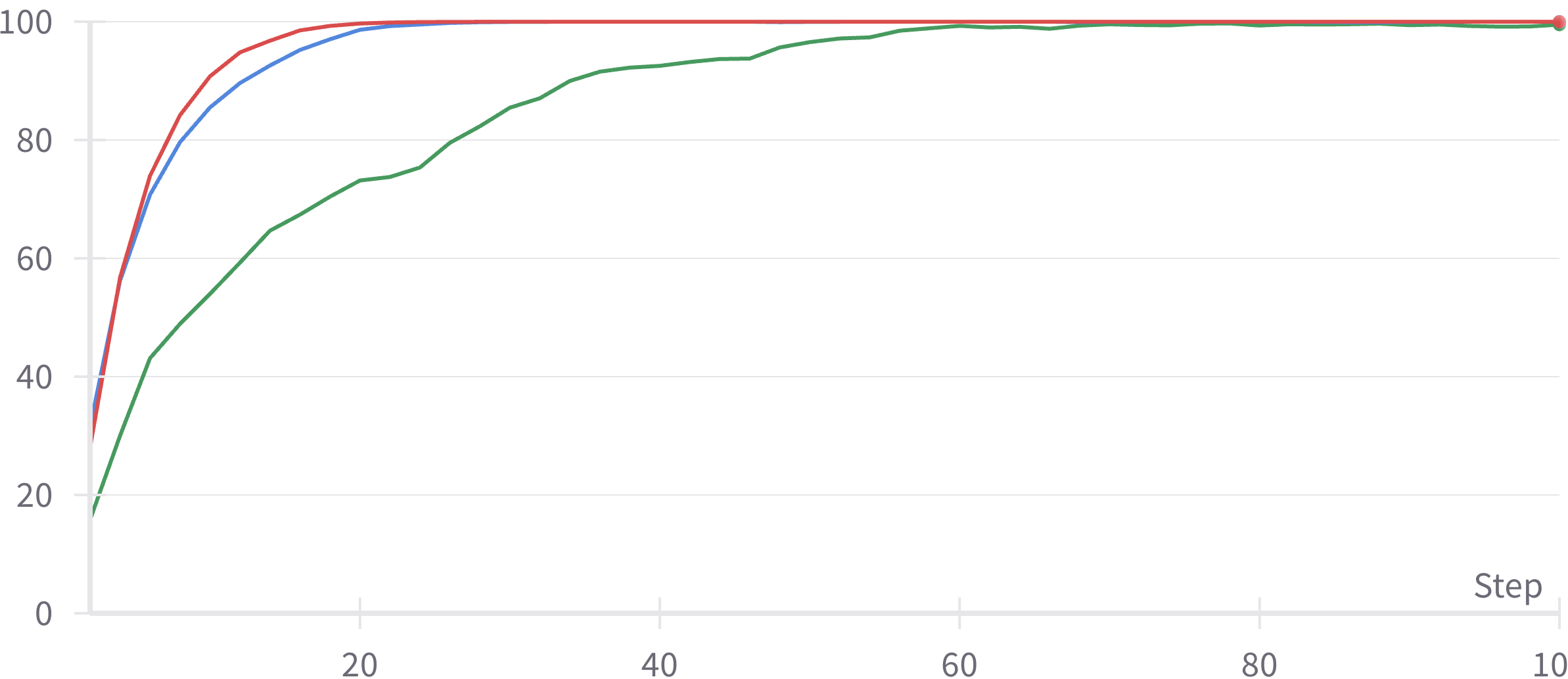}
        \caption{Plant Village}
        \label{fig:sub5}
    \end{subfigure}
    \hfill
    \begin{subfigure}[htbp]{0.3\textwidth}
        \centering
        \includegraphics[width=\textwidth]{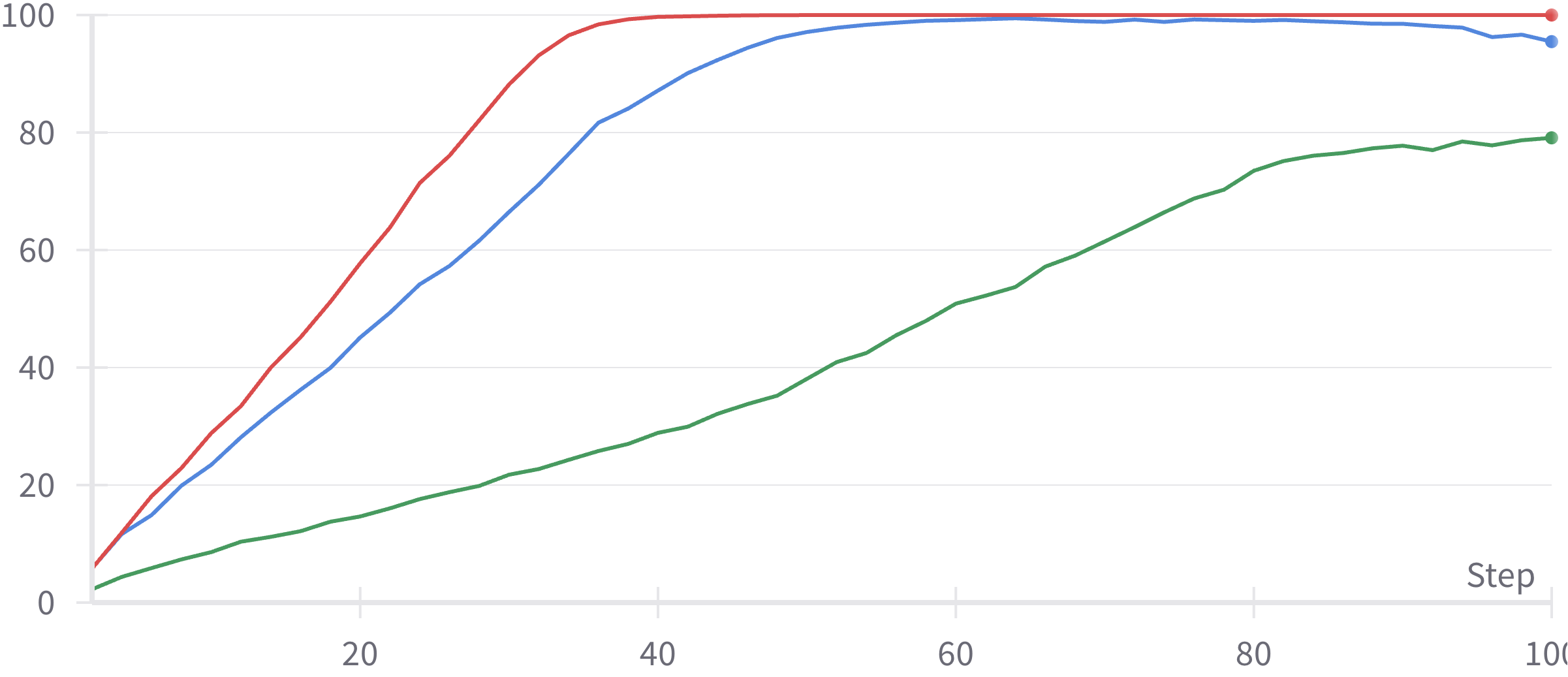}
        \caption{Insects}
        \label{fig:sub6}
    \end{subfigure}
    \vspace{5mm}

    \caption{Training curves for 6 different datasets. Multi-dataset transfer (red) converges fastest, but also OT-Transfer (blue) converges much faster than training SmoothDARTS from scratch (green). The y-axis is the top-1 training accuracy and the x-axis is the number of training epochs.}
    \label{fig:composite}
\end{figure*}

\subsection{Effect of multiple datasets for supernet pretraining}
We now compare running SmoothDARTS from scratch versus OT-Transfer and Multi-Dataset Transfer, in which the latter is evaluated in a leave-one-out fashion, pretraining a supernet on all datasets except the one held out as the target dataset. The top-1 accuracy results are shown in Table \ref{tab:lootable}    where the \textit{relative performance} of both OT Transfer and the Oracle(The dataset with most optimal transfer) versus the models designed from scratch, such that $RI_{OT} = \frac{ACC_{OT} - ACC_{scratch}}{ACC_{scratch}} $ and $ RI_{Oracle} = \frac{ACC_{Oracle} - ACC_{scratch}}{ACC_{scratch}} $. We observe that supernet transfer using Optimal Transport is effective, often outperforming training DARTS from scratch, and often (but not always) close to the Oracle selection. Only on 7 out of 27 datasets, the difference with the Oracle is greater than 5 percent. We observe that Multi-Dataset Transfer significantly outperforms \textit{both} OT-Transfer and the Oracle almost every time. Hence, very large and diverse pretraining datasets almost always guarantee the best transfer performance.

% \begin{figure*}[ht]
%     \centering
%     \includegraphics[width=\textwidth]{figures/loo.png}
%     \caption{Transfer learning from all datasets to the target dataset in a leave-one-out setting}
%     \label{fig:transfer_leave_one_out}
% \end{figure*}

\subsection{Convergence speed of supernet transfer}
Figure \ref{fig:composite} shows the learning curves from 6 representative datasets. We observe that both OT-Transfer and Multi-Dataset Transfer converge much faster than training from scratch, about 3 to 5 times faster, and even find better models than running SmoothDARTS from scratch for 100 epochs, with performance gaps ranging from a few percent up to 20 percent. This suggests that there is information in the source datasets that cannot be learned from the target dataset alone.

\subsection{Analysis of the final architectures after discretization}
\begin{figure}[h]
    \centering
    \includegraphics[width=0.8\columnwidth]{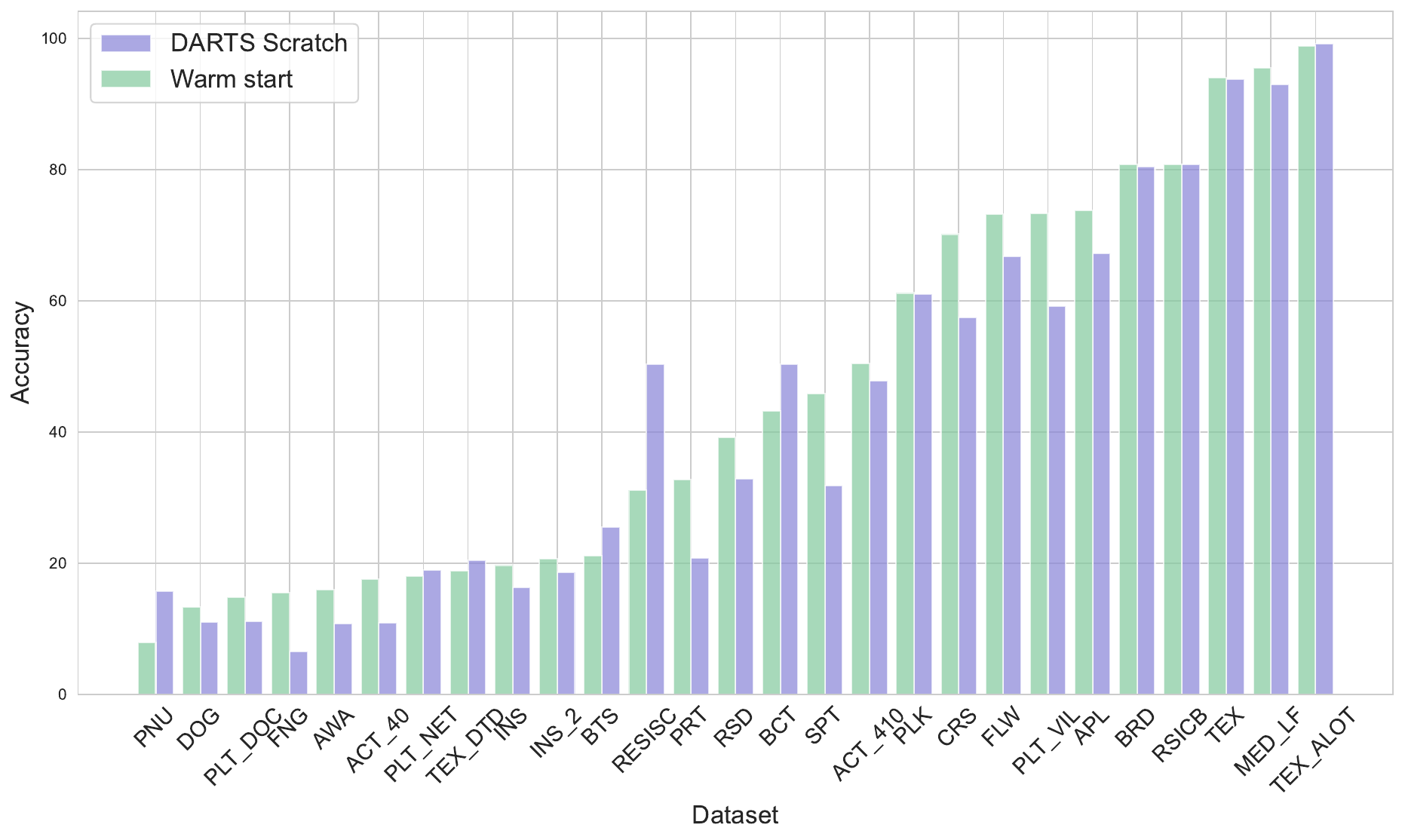}
    \caption{Performance of final obtained architectures, (We use Meta-album tags here for better readability)}
    \label{fig:final-archs}
\end{figure}

Finally, we compared the final architectures obtained from training from scratch and from supernet transfer learning from a single source dataset(i.e after discretization). Figure \ref{fig:final-archs} shows that on 10 out of 27 datasets, the performance gains from warm-starting are greater than 5 percent. On 3 datasets, training from scratch yielded better performance, indicating that OT distances did not select the right source dataset, which is interesting for further research.

\section{Conclusion and Discussion}
In this work, we demonstrated that pretraining and transferring supernets is a promising approach for both NAS and transfer learning, making NAS converge significantly faster and yielding better models than those discovered by state-of-the-art NAS techniques. Considering that supernet training is the most expensive step in DARTS-like frameworks, this work provides practical solutions and useful insights for future work on transfer learning in NAS.

We observe that negative and positive transfer does exist in SmoothDARTS and hence there should be a transferability metric that can select the right pretrained supernets. We show that Optimal Transport is a promising framework to achieve this, yielding positive transfer with almost every target dataset.
Moreover, we observe that pretraining supernets on large mixtures of source datasets leads to very robust transfer learning performance, both in accuracy and convergence speed. 

We also find that our results differ from earlier studies such as \citet{nasnanlysisi}, and can provide several reasons for these differences. First, ~\citet{nasnanlysisi} focuses on ENAS~\citep{enas} and the original DARTS implementation, which are demonstrated to be very unstable. ~\citet{enas-problem} have shown that ENAS performance stems from good search space design rather than efficient search, and the controller doesn't seem to learn anything. Several studies have also pointed out the instability of the first DARTS implementation as described by ~\citet{robustdarts}. Second, we use SmoothDARTS~\citep{smoothdarts}, which makes DARTS training much more stable and provides more generalizable results, as well as Meta-Album, which is a much more complex and rich collection of datasets compared to IMAGENET22K, which allows us to study transfer learning much more effectively.

\par

In conclusion, our work contributes the following observations and practical tools for efficient supernet transfer learning:
\begin{enumerate}
    \item Pretraining supernets and using them to warm-start NAS is an effective way to speed up NAS and even discover better models. These supernets are trained very often but almost always discarded in NAS. Instead, they can be stored and used to make NAS more accessible.
    
    \item Based on extensive experiments, we find that supernet transfer speeds up the convergence of differentiable NAS methods by 3x-5x. This finding shows that machine learning engineering techniques like \textbf{early stopping} can be used in supernets.
    
    \item Using a transferability metric based on optimal transport provides a robust and effective way to select pretrained transfer learning for DARTS-like frameworks.
    
    \item Pretraining supernets on large and diverse mixtures of datasets is significantly more robust and transfers much better than pretraining on a single dataset. As such, creating these pretrained supernets and making them available can be a boon for both NAS and transfer learning in general.

\end{enumerate}
    Compared to traditional transfer learning, it allows us to finetune not only the model weights, but also the model architecture, allowing optimization to novel objectives or architecture constraints.(We also aim to provide a link to our model zoo of pretrained supernets.)

\section{Limitations}
The OT-based distance metric can be hard to scale if the dataset pool is very large, as the dataset similarity computation increases linearly with the number of prior datasets, this limitation can be overcame with using linear approximation of optimal transport approaches. Secondly, our work assumes that a supernet-like structure has been defined for the model family that we want to pretrain supernets for. This limitation only applies in a  setting where one does not have access to pretrained supernets. 

\section{Future Work}
Selecting the best dataset mixtures to pretrain supernets is an obvious next challenge. This would lead to optimally pre-trained supernets that anyone can simply download and fine-tune to new tasks (without having to use model zoos). We also aim to explore the supernet pretraining with transformer models instead of CNNs, to add capacity and allow for larger-scale pretraining. This could be done using NASViT~\citep{nasvit} supernets or Once-For-All~\citep{Cai2020Once-for-All:} models.

Another direction of future work would be \textit{source-free} transfer learning metrics on DARTS supernet where one does not have access to source datasets but does have access to pretrained supernets.  

Finally, this work focuses on fast and efficient transfer learning, but doesn't yet address \textit{continual learning} where also catastrophic forgetting is important, next to fast adaptation. Yet, there are obvious avenues of research for different forms of continual learning. Regularization-based continual learning techniques could be included to fine-tune towards both fast adoption and minimal forgetting, for instance by freezing certain parts of the network or using meta-learned optimizers. Replay-based techniques could be used in a straightforward way to finetune the supernets with experience replay~\citep{Rolnick2018ExperienceRF}, mixing in examples from previous tasks. Finally, this method could be developed further into architecture-based continual learning techniques that also scale well, since we can continually adapt the architecture (e.g., continually learning the architecture weights) instead of simply extending the network with new modules. 

We hope that supernet transfer will inspire much more work in these directions.

\vfill\null 

\bibliography{main}
\bibliographystyle{collas2025_conference}

\appendix
\newpage
\section{Additional Experiments and analysis}
 We also perform a correlation analysis of the OT score with the final performance of the supernet itself in Table \ref{tab:top_10_kendall_tau_values}. The scores shows a mild correlation between dataset similarities and the training of supernets. This does open more work in the area of transferability metrics for supernets additional to the proposed method.

\begin{table}[h!]
\centering
\begin{tabular}{|l|c|c|c|}
\hline
\textbf{Dataset} & \textbf{Kendall} & \textbf{Weighted Kendall} & \textbf{Pearson}  \\
\hline
SPT      & 0.480 & 0.485 & 0.412 \\
BTS      & 0.476 & 0.374 & 0.475  \\
ACT\_410 & 0.382 & 0.362 & 0.492  \\
ACT\_40  & 0.377 & 0.578 & 0.353  \\
FNG      & 0.349 & 0.505 & 0.418 \\
AWA      & 0.331 & 0.482 & 0.345 \\
PLT\_NET & 0.261 & 0.447 & 0.269  \\
RSICB    & 0.259 & 0.241 & 0.589  \\
PLT\_DOC & 0.159 & 0.367 & 0.351  \\
INS\_2   & 0.192 & 0.145 & 0.368  \\
\hline
\end{tabular}
\caption{Top 10 datasets based on Kendall tau values using the OTDD metric.}
\label{tab:top_10_kendall_tau_values}
\end{table}

\section{Experimental analysis }
All the experiments were performed on NVIDIA A8000 GPU. The code is provided in the zip file with accompanying instructions. We also provide performances of all the experiments in the CSV folder. We use meta-album dataset, the number of classes, image domains and dataset information can be found on \url{https://meta-album.github.io/} \\
\subsection{Runtime}
We trained the supernet for 50 epochs with S1 search space from SmoothDarts with random perturbation. The runtime for supernet training was not affected with and without warm starting though we notice with training curves that the convergence of supernet happens much earlier in warm starting than training from scratch. For final architectures we observe the training time is 10 percent lower for warm starting than without warm starting which emperically suggests that warm starting can also lead to more efficient final architectues but more investigation on this topic is required.

\subsection{Final architecture experiments}
\begin{figure*}[!ht]
    \centering
    % First row with three subfigures
    \begin{subfigure}[htbp]{\textwidth}
        \centering
        \includegraphics[width=\textwidth]{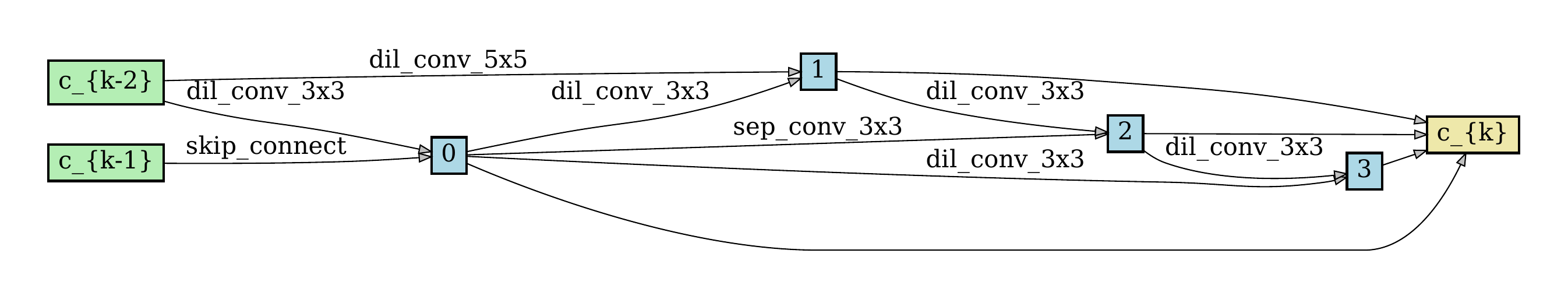}
        \caption{PLK Normal}
        \label{fig:sub1}
    \end{subfigure}
    \hfill
    \begin{subfigure}[htbp]{\textwidth}
        \centering
        \includegraphics[width=\textwidth]{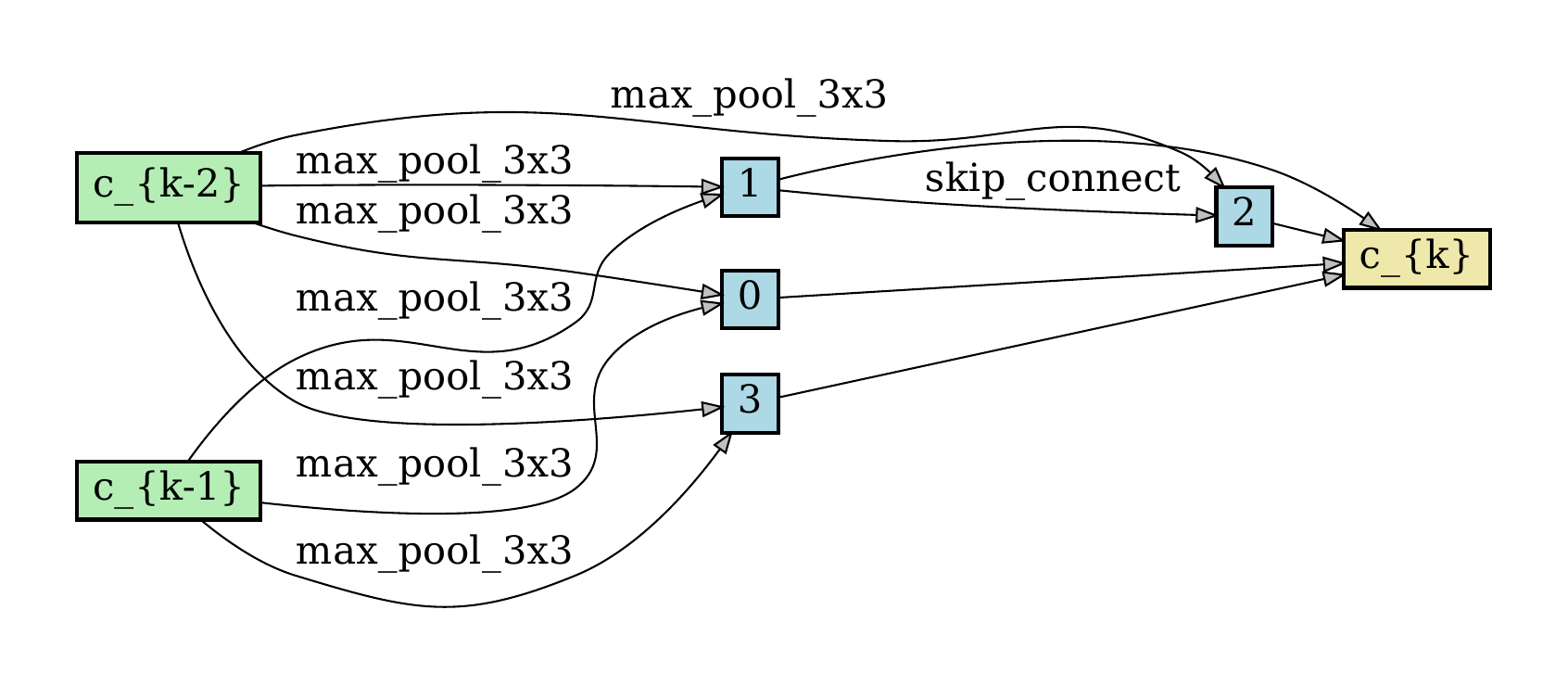}
        \caption{PLK reduction}
        \label{fig:sub2}
    \end{subfigure}
    \hfill
    \vspace{5mm}
\caption{Normal and reduction cell for PLK dataset with training from scratch}
\label{plknws}
\end{figure*}

We compared final architectures obtained by warm-started supernet vs the supernet trained from scratch. We use a batch size of 32 and a layer size of 30 for the final architectures. We report the performances of these final architectures and also visualize the normal and reduction cell for the PLK dataset in Figure \ref{plknws} and Figure \ref{plkws}.

\begin{figure*}
    \centering
    \includegraphics[width=\linewidth]{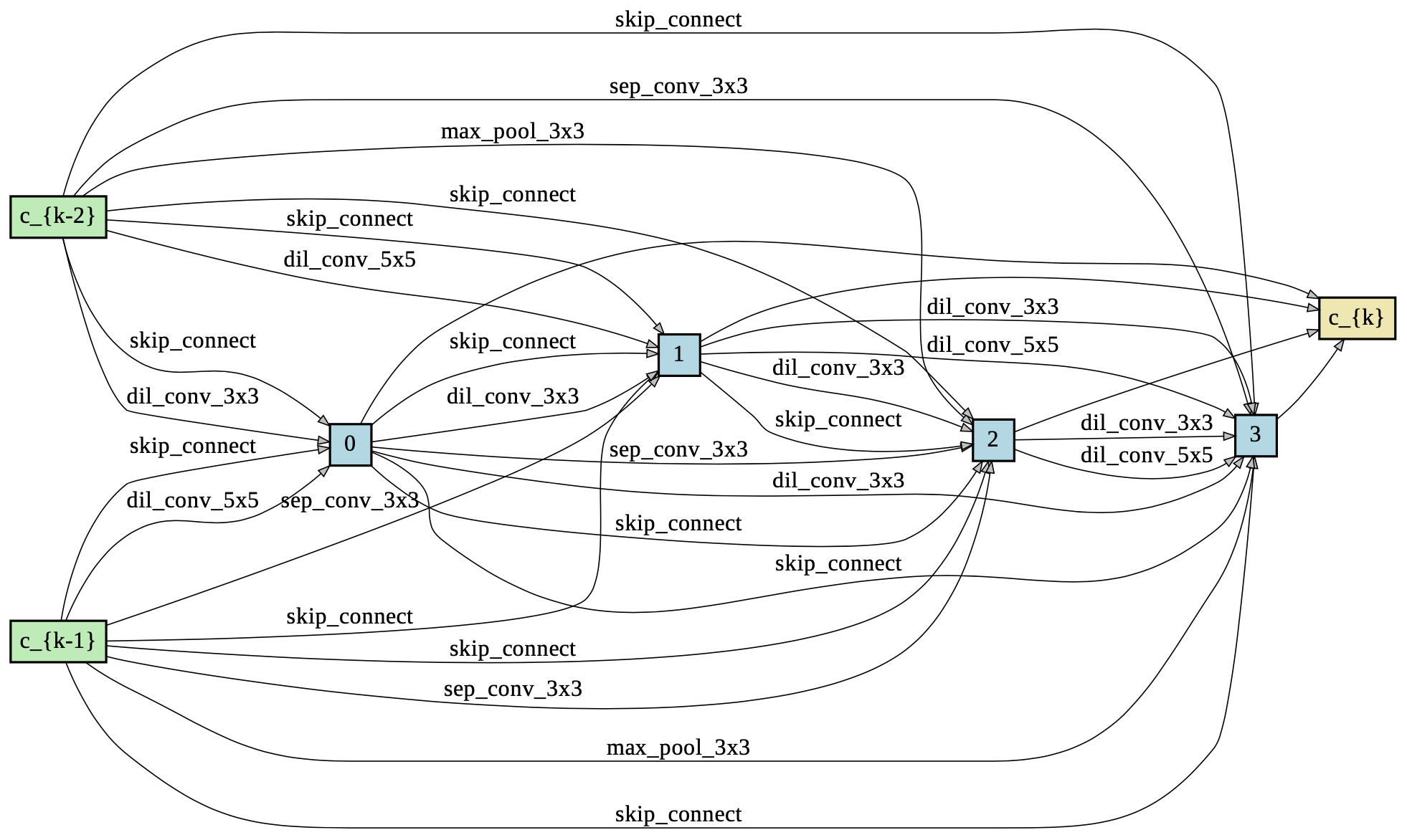}
    \caption{Normal cell search space from SmoothDARTS S1 used in our experiments}
    \label{fig:enter-label}
\end{figure*}
\begin{figure*}
    \centering
    \includegraphics[width=\linewidth]{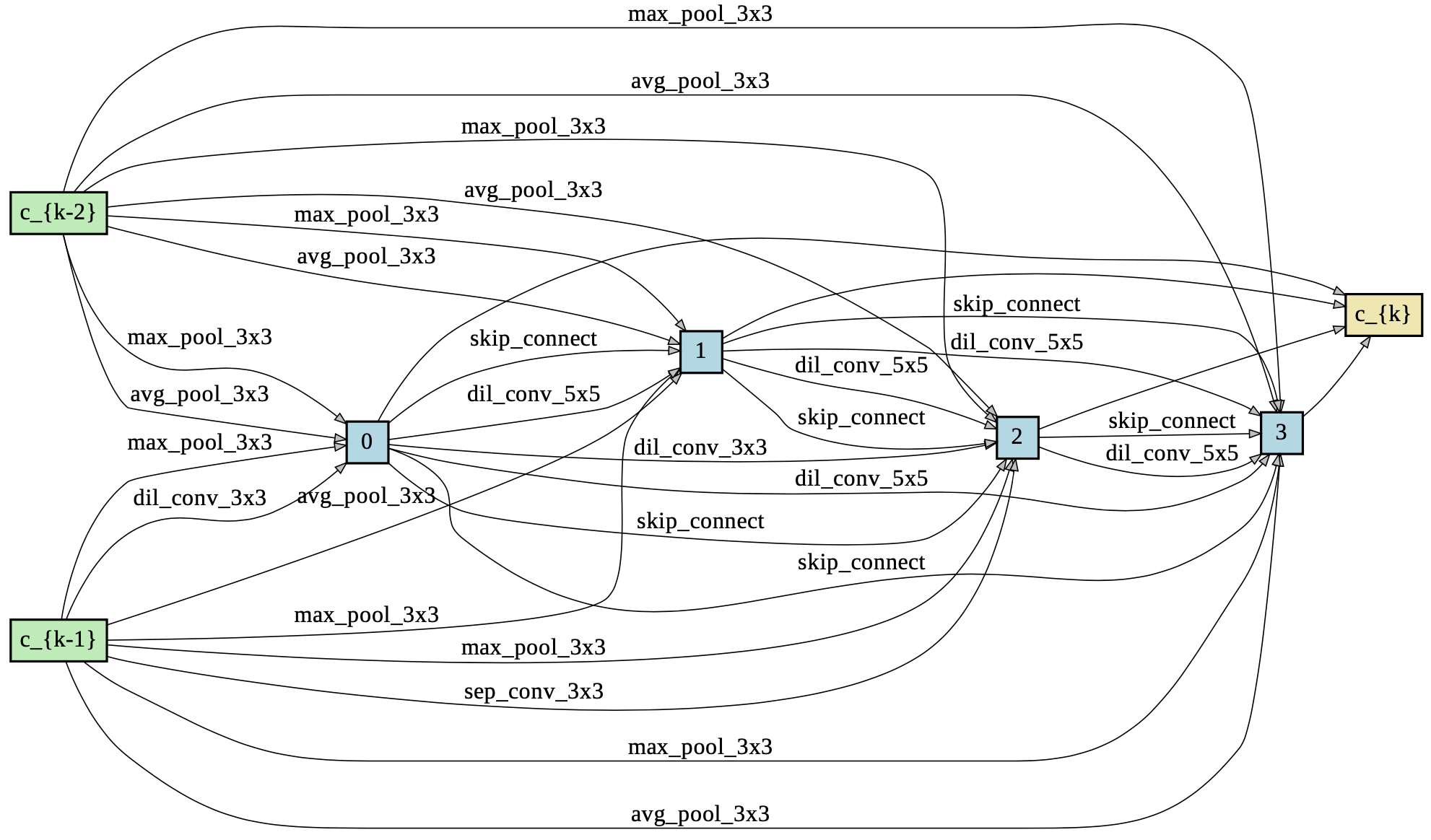}
    \caption{Reduction cell search space from SmoothDARTS S1 used in our experiments}
    \label{fig:enter-label}
\end{figure*}
\begin{figure*}[!ht]
    \centering
    % First row with three subfigures
    \begin{subfigure}[htbp]{0.6\textwidth}
        \centering
        \includegraphics[width=\textwidth]{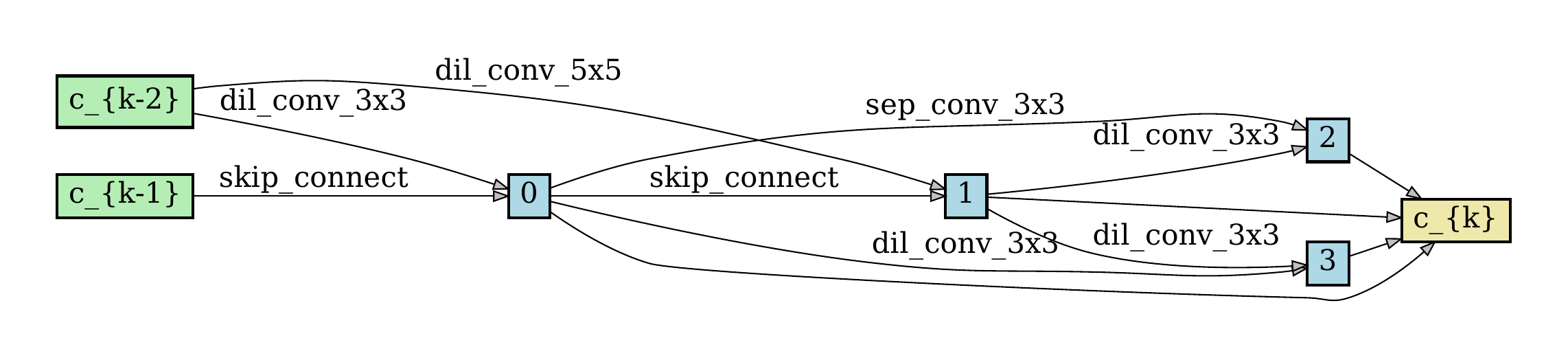}
        \caption{PLK\_WS Normal}
        \label{fig:sub1}
    \end{subfigure}
    \hfill
    \begin{subfigure}[htbp]{0.6\textwidth}
        \centering
        \includegraphics[width=\textwidth]{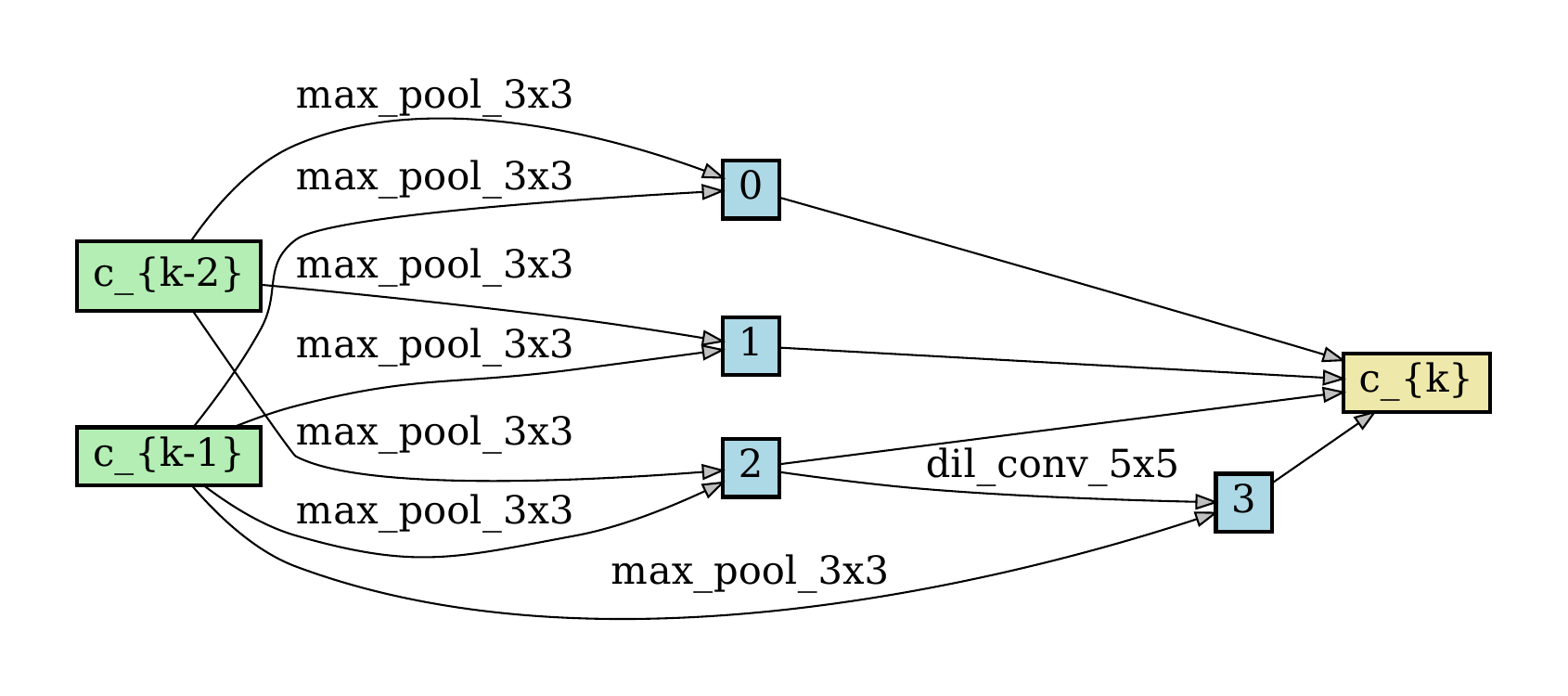}
        \caption{PLK\_WS reduction}
        \label{fig:sub2}
    \end{subfigure}
    \hfill
    \vspace{5mm}
\caption{Normal and reduction cell for PLK dataset with warm starting}
\label{plkws}
\end{figure*}
\section{OT Similarity matrix}

A heatmap of the OT similarities is shown in Figure \ref{fig: 
 heatmap}.\label{sec:rationale}
Note that this algorithm allows for flexibility in the choice the embedding function and OT distance metric since these can depend on the type of data one is dealing with. In the remainder of this paper, we'll evaluate this method for image classification problems. We use Resnet18~\citep{resnet}, pretrained on ImageNet as our embedding function $\phi$, and use entropic regularisation of 1e-1 with 1000 samples. Since we do classification, we select OTDD~\citep{otdd} as our OT distance metric as it incorporates label cost as well.
\begin{figure}
    \centering
    \includegraphics[width=0.8\columnwidth]{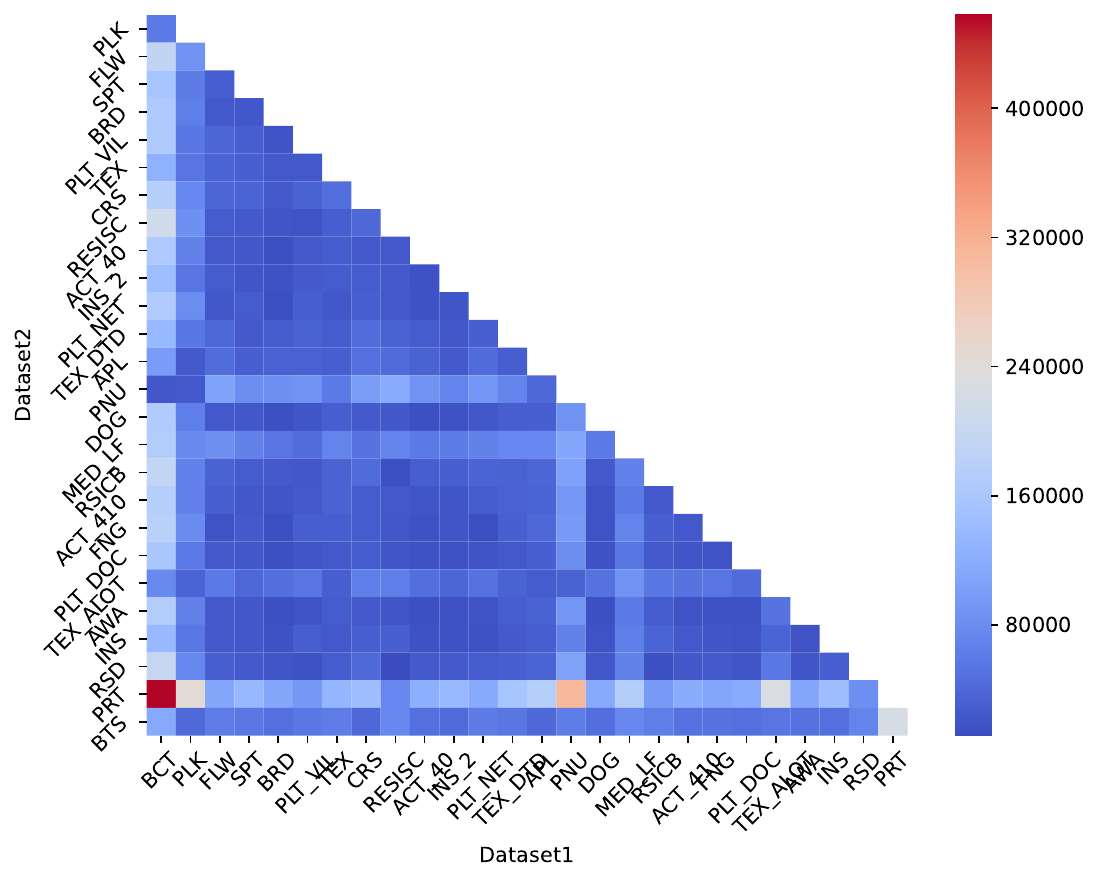}
    \caption{Heatmap of dataset similarities calculated via $d_{ot}$}
    \label{fig: heatmap}
    \vspace{2mm}
\end{figure}

\begin{figure}
    \centering
    \includegraphics[width=0.7\columnwidth]{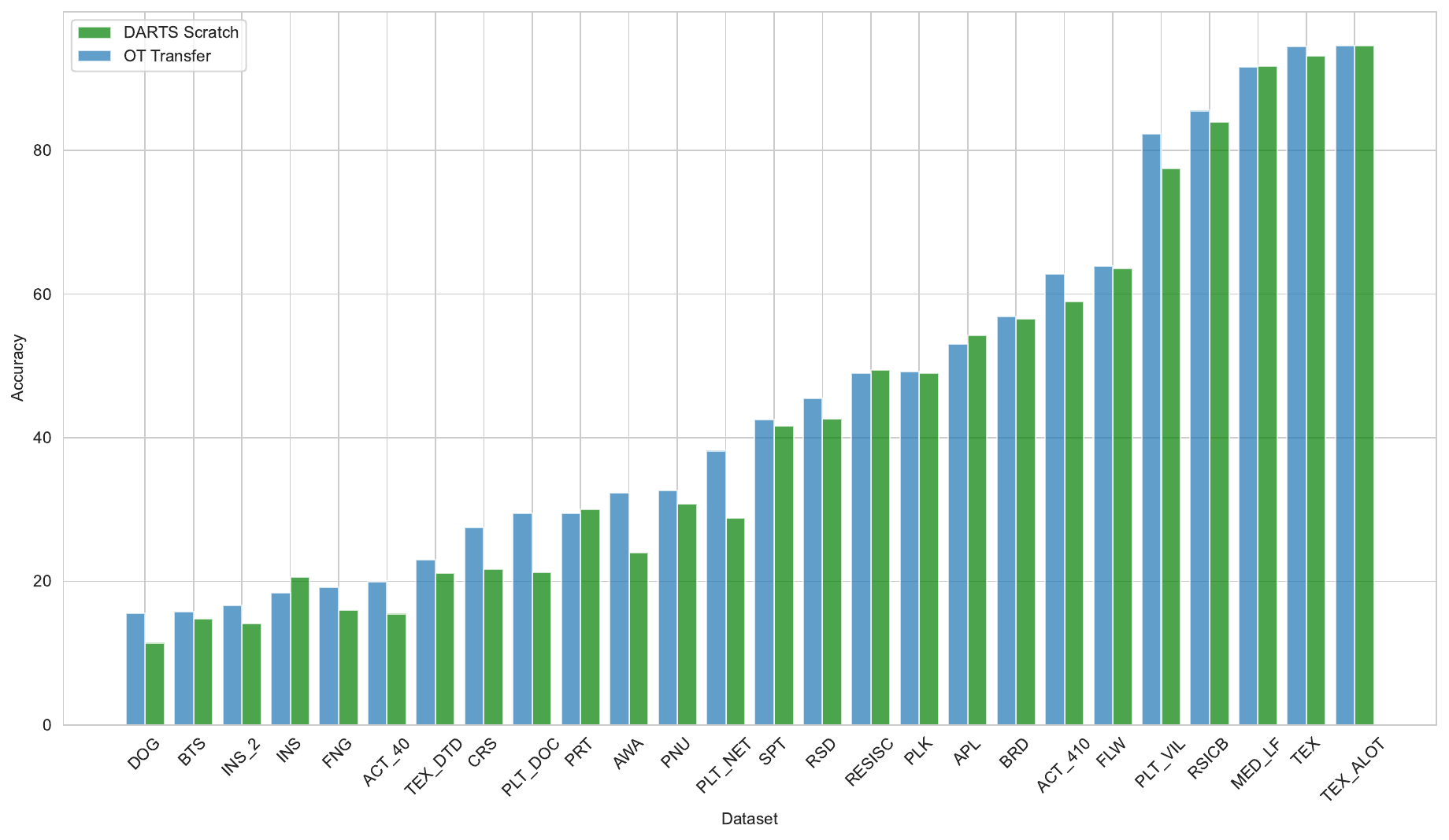}
    \caption{Comparison of model performance between supernet transfer with OT-based distances vs training SmoothDARTS from scratch. (We use Meta-album tags here for better readability)}
    \label{fig:transfer_otdd}
\end{figure}

\end{document}